\documentclass[twoside,11pt]{article}
\usepackage{jmlr2e}
\usepackage[latin9]{inputenc}
\usepackage{amsmath}
\usepackage{amssymb}
\usepackage{graphicx}
\usepackage{comment}
\usepackage{enumerate}
\usepackage{algorithm}
\usepackage{algorithmic}
\DeclareGraphicsExtensions{.pdf,.jpg,.png}

\usepackage{caption}
\usepackage{lipsum}
\usepackage{xfrac}
\usepackage{array}
\usepackage{url}
\RequirePackage{pdflscape}
\newcolumntype{M}[1]{>{\centering\arraybackslash}m{#1}}
\newcolumntype{N}{@{}m{0pt}@{}}

\newcommand{\argmin}{\mathrm{arg}\displaystyle\min}

\ShortHeadings{Cross-Validated Variable Selection in Tree-Based Methods}{Painsky and Rosset} \firstpageno{1}

\begin{document}

\title{Cross-Validated Variable Selection in Tree-Based Methods Improves Predictive Performance}

\author{\name Amichai Painsky \email amichaip@eng.tau.ac.il \\
       \addr School of Mathematical Sciences\\
       Tel Aviv University\\
       Tel Aviv, Israel
       \AND
       \name Saharon Rosset \email saharon@post.tau.ac.il \\
       \addr School of Mathematical Sciences\\
       Tel Aviv University\\
       Tel Aviv, Israel}

\editor{}

\maketitle

\begin{abstract}
Recursive partitioning approaches producing tree-like models are a long standing staple of predictive modeling, in the
last decade mostly as ``sub-learners'' within state of the art
ensemble methods like Boosting and Random Forest. However, a
fundamental flaw in the partitioning (or splitting) rule of
commonly used tree building methods precludes them from treating
different types of variables equally. This most clearly manifests in
these methods' inability to properly utilize categorical variables
with a large number of categories, which are ubiquitous in the new
age of big data. Such variables can often be very informative, but
current tree methods essentially leave us a choice of either not
using them, or exposing our models to severe overfitting. We propose a conceptual framework to splitting using leave-one-out (LOO) cross validation for selecting the splitting variable, then performing a regular split (in our case, following CART's approach) for the selected
variable. The most important consequence of our approach is that
categorical variables with many categories can be safely used in
tree building and are only chosen if they contribute to
predictive power. We demonstrate in extensive simulation and real
data analysis that our novel splitting approach significantly
improves the performance of both single tree models and ensemble
methods that utilize trees. Importantly, we design an algorithm for
LOO splitting variable selection which under reasonable assumptions does not increase the overall
computational complexity compared to CART for two-class classification. For regression tasks, our approach carries an increased computational burden, replacing a $O(log(n))$ factor in CART splitting rule search with an $O(n)$ term.
\end{abstract}


\section{Introduction}
\begin{figure}[!ht]
\begin{center}
\includegraphics[width = 0.45\textwidth,bb= 90 95 705 600,clip]{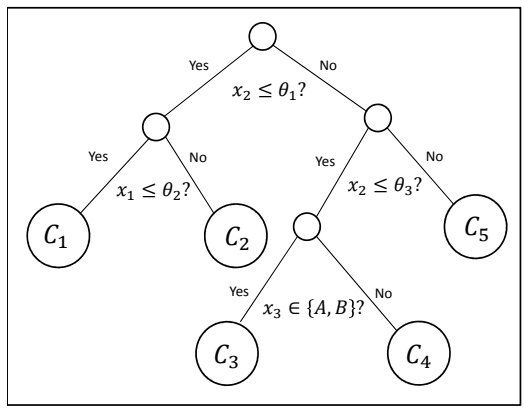}
\caption{A tree-based model example. $x_1, x_2$ are numerical/ordinal features and $x_3$ is a categorical feature}

\label{fig:tree}
\end{center}
\end{figure}

The use of trees in predictive modeling has a long history, dating
back to early versions like CHAID \citep{kass1980exploratory}, and gaining importance in the 80s and 90s through the introduction of approaches like CART \citep{breiman1984classification} and C4.5/C5.0 \citep{quinlan2014c4, quinlan2004data}. A tree describes a recursive
partitioning of a feature space into rectangular regions intended to
capture the relationships between a collection of explanatory
variables (or features) and a response variable, as illustrated in
Figure \ref{fig:tree}. The advantages of trees that made them attractive include
natural handling of different types of features
(categorical/ordinal/numerical); a variety of approaches for dealing
with missing feature values; a natural ability to capture non-linear
and non-additive relationships; and a perceived interpretability and
intuitive appeal. Over the years it has become widely understood
that these advantages are sometimes overstated, and that tree models
suffer from a significant drawback in the form of inferior
predictive performance compared to modern alternatives (which is not
surprising, given the greedy nature of their model building
approaches). However the last 15 years have seen a resurgence in the
interest in trees as ``sub-learners'' in ensemble learning
approaches like Boosting \citep{friedman2001} and Random Forest \citep{breiman2001random}. These approaches take advantage of the favorable properties described above, and mitigate the low accuracy
by averaging or adaptively adding together many trees. They are
widely considered to be among the state of the art tools for
predictive modeling \citep{hastie2009elements}.

As mentioned, popular tree building algorithms like CART can handle
both numerical and categorical features and build models for
regression, two-class classification and multi-class classification.
The splitting decisions in these algorithms are based on optimizing
a splitting criterion over all possible splits on all variables. The
split selection problems that arise present both computational and
statistical challenges, in particular for categorical features with
a large number of categories. Assume $K$ categories overall, then
the space of possible binary splits includes $O(2^K)$ candidates,
and a naive enumeration may not be feasible. However, as
demonstrated in CART \citep{breiman1984classification}, there is no need to enumerate over all
of them in order to find the optimal one for regression or two-class
classification, because it is enough to sort the categories by their
mean response value and consider only splits along this sequence.
The complexity of splitting is therefore $O(\max(K \log(K), n))$,
where $n$ is the number of observations being split. However, if $K$
is large, the splitting still presents a major statistical
(overfitting) concern. At the extreme, if $K=n$ (for example, if the
categorical variable is in fact a unique identifier for each
observation), then it is easy to see that a single split on this
variable can perfectly separate a two-class classification training
set into its two classes. Even if $K$ is smaller than $n$ but large,
it is intuitively clear (and demonstrated below) that splitting on
such a categorical variable can result in severe overfitting. This
is exasperated by the fact that this overfitting in fact causes
common tree building splitting criteria like Gini to preferably
select such categorical variables and ``enjoy'' their overfitting.

Some popular tree building software packages recognize this problem
and limit the number of categories that can be used in model
building. In {\sf R}, many tree-based models limit the number of categories to $32$ (as {\sf randomForest}, for example), while Matlab's regression-tree and classification-tree routines ({\sf fitrtree, fitctree}) leave this parameter for the users to define. In what follows we denote implementations which discard categorical features with a number of categories larger than $K$ as \textit{limited-$K$} while versions which do not apply this mechanism will be denoted as \textit{unlimited-$K$}. 
The ad-hoc limited-$K$ approach is of course far from satisfactory, because the chosen number can be
too small or too large, depending on the nature of the specific
dataset and the categorical features it contains. It also fails to
address the conceptual/theoretical problem: since features that
are numerical or ordinal or have a small number of categories present
a smaller number of unique splits than categorical features with
many categories, there is a lack of uniformity in split criteria
selection. Finally, some large datasets may well have categorical
features with many categories that are still important for good
prediction, and discarding them is counterproductive.

In this work we amend these concerns through a simple modification:
We propose to select the variable for splitting based on LOO scores
rather than on training sample performance, an approach we term 
Adaptive LOO Feature selection (ALOOF). A naive implementation
of ALOOF approach would call for building $n$ different best splits on each
feature, each time leaving one observation out, and select the
feature that scores best in LOO. As we demonstrate below, this
implementation can be avoided in most cases. The result of our
amendment is a ``fair'' comparison between features in selection,
and consequently the ability to accommodate all categorical features
in splitting, regardless of their number of categories, with no
concern of major overfitting. Importantly, this means that truly
useful categorical variables can always be taken advantage of.

\begin{figure}[!ht]
\centering
\includegraphics[width = 0.5\textwidth,bb= 40 185 580 590,clip]{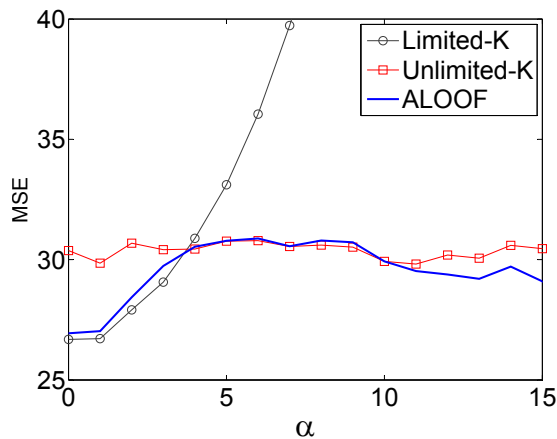}
\caption{Predictive performance of the two CART versions and ALOOF for the model with one categorical and one numerical feature described in the text. As can be seen, ALOOF successfully identifies the better approach throughout the range of $\alpha$ (which determines the strength of the signal from the categorical variable).}
\label{fig:sim3}
\end{figure}

The preceding concepts are demonstrated in Figure \ref{fig:sim3}. The simulation setting includes $n=300$ observations and an interaction between a numerical variable $x_1$ and a categorical variable $x_2$ with $K=50$ categories denoted $c_1,...,c_{50}$, with a parameter $\alpha$ controlling the strength of the interaction:
\begin{eqnarray}
\label{intro_model}
 y &=& \alpha \cdot \left( {\mathbb I}\left[x_1>0\right] {\mathbb I}\left[ x_2 \in \{c_1,...,c_{25}\} \right]\right. + \left. {\mathbb I}\left[x_1 \leq 0\right] {\mathbb I}\left[ x_2 \in \{c_{26},...,c_{50}\} \right] \right) + \epsilon, 
\end{eqnarray}
where $\epsilon \sim N(0,1)$ i.i.d. At small $\alpha$ the signal from the categorical variable is weak and less informative. In this case, the limited-$K$ approach is preferable (here, as in all subsequent analyses, we use a CART limited-$K$ Matlab implementation that eliminates categorical variables with $K>32$ values). As $\alpha$ increases, the categorical variable becomes more informative and unlimited-$K$ (which refers to a CART implementation on Matlab that does not discard any categorical variable) outperforms the limited-$K$ approach for the same $n$ and $K$. ALOOF (which is also implemented on a Matlab platform) successfully tracks the preferred approach in all situations, even improving slightly on the unlimited-$K$ approach for large $\alpha$ by selecting the categorical variable less often in the tree.

The rest of this paper is organized as follows: In Section \ref{Sec2} we
formalize the LOO splitting approach in ALOOF and discuss its statistical
properties, we then describe our algorithmic modifications for
implementing it in Section \ref{Sec3} and demonstrate that for two-class
classification they lead to an approach whose computational
complexity is comparable to CART splitting under reasonable
assumptions. In Section \ref{Sec4} we demonstrate via some toy simulations
the striking effect that ALOOF can have on improving the
performance of tree methods. We begin Section \ref{Sec5} with a comprehensive case study, where large-K categorical variables are prevalent and important. We demonstrate the significant improvement in both interpretability and prediction from using ALOOF vs regular CART splitting. We then preset an extensive real data study in both regression and classification, using both single CART trees and ensemble methods, where ALOOF offers across-the-board improvement.

\subsection{Related work} We focus here on two threads of related
work, which relates to our approach from two different aspects.
The Conditional Inference Trees (CIT) framework \citep{hothorn2006unbiased} proposes a method for unbiased variable selection through conditional permutation tests. At each node, the CIT procedure tests the global null hypothesis of complete independence between the response and each feature, and selects the feature with the most significant association (smallest p-value). It terminates once the global null can no longer be rejected. The advantage of this approach is that it proposes a unified framework for recursive partitioning which embeds tree-structured models into a well defined statistical theory of conditional inference procedures. However, practically it requires the computation of p-values under permutations tests, either by exact calculation, Monte Carlo simulation, or asymptotic approximation. These permutation tests are typically computationally expensive and necessitate impractical run-time as the number of observations grows (as demonstrated in our experiments).
More critically, despite its well established theory, the statistical significance criterion the CIT procedure uses is not the relevant criterion for variable selection in tree-based modeling. Notice that in each node, the feature that should be most ``informative" is the one that splits the observations so that the (test-set) generalization error is minimal. This criterion does not typically select the same variable that results with the smallest p-value under the null assumption on the train-set.
As our experiments in Section \ref{Sec5}  show, our ALOOF method, based on the generalization error criterion, is superior the the CIT approach, where the CIT is computationally feasible.

In a second line of work, Sabato and Shalev-Schwartz \citep{sabato2008ranking} suggest ranking categorical features according to their generalization error. However, their approach is limited to considering K-way splits (i.e., a leaf for every categorical value), as opposed to binary splits which group categories,  which characterize CART and other tree methods commonly used in statistics. 
Their appraoch, which is also solely applicable for classification trees, offers strong theoretical results and low computational complexity (as it does not perform actual splitting), albeit in the K-way scenario which is significantly simpler theoretically than the binary split case we are considering, which requires optimizing over $O(2^K)$ possible splits. Thus our goals are similar to theirs, but since we are dealing with a more complex problem we cannot adopt either their computational approach or their theoretical results. In our classification examples we compare our approach to theirs, demonstrating the advantage of binary splitting for attaining better predictive performance in practice.  The approach of \cite{sabato2008ranking} subsumes and improves on various previous lines of work in the machine learning literature on using cross validation to select variable in K-way splitting scenarios \citep{Frank1,Frank2}.

\section{Formulation of ALOOF} \label{Sec2}
We present here the standard CART splitting approach \citep{breiman1984classification} based on least
squares for regression and Gini index (closely related to least
squares) for classification, and then describe our modifications to
adapt them to LOO splitting variable selection. We choose to
concentrate on CART as arguably the most widely used tree
implementation, but our LOO approach can be adapted to other
algorithms as well. 

\subsection{CART splitting rules}

Assume we have $n$ observations denoted by their indexes $\{1,...,n\}$. Each observation $i$ is comprised of the pair $(x_i,y_i)$, where  $x_i$ is a $p$ dimensional vector containing $p$ candidate variables to split on. We assume WLOG that the variables $x_{\cdot 1}...x_{\cdot q}$ are categorical with number of categories $K_1,...,K_q$ respectively, and the features $x_{\cdot q+1},...,x_{\cdot p}$ are numerical or ordinal (treated identically by CART).  
A split $s$ is a partition of the $n$ observations into two subsets $R(s),L(s)$ such that $R(s)\cap L(s)=\emptyset$ and $R(s)\cup L(s) = \{1,...,n\}$. 
For each variable $j$, denote the set of possible splits by $S_j$ and their number by $s_j = |S_j|$. 
Categorical variable $j$ has $s_j = 2^{K_j-1}-1$ possible binary splits, and each numerical/ordinal variable has $s_j \leq n$ possible splits between its unique sorted values. 

A specific split comprises a  selection of a splitting variable $j$ and a specific split $s \in S_j$, and is evaluated by the minimizer of an ``impurity'' splitting criterion given the split $s$, denoted by ${\cal L}(s)$. 

In regression, CART uses the squared error loss impurity criterion:
$$ {\cal L}(s) = \sum_{i \in L(s)} (y_i - \bar{y}_L)^2 +  \sum_{i \in R(s)} (y_i - \bar{y}_R)^2,$$
where $\bar{y}_L, \bar{y}_R$ are the means of the response $y$ over the sets $L(s), R(s)$ resepectively. 

In two-class classification, it uses the Gini index of impurity:
$$ {\cal L}(s) = n_L \hat{p}_L (1-\hat{p}_L) +  n_R  \hat{p}_R (1-\hat{p}_R), $$
where $n_L,n_R$ are the numbers of observations in $L(s),R(S)$ respectively and $\hat{p}_L,\hat{p}_R$ are the observed proportions of ``class 1'' in $L(s),R(S)$. 
The Gini index is easily seen to be closely related to the squared error loss criterion, as it can be viewed as the cross-term of a squared error loss criterion with 0-1 coding for the classes. 
The Gini criterion for multi-class classification is a straight forward generalization of this formula, but we omit it since we do not consider the multi-class case in the remainder of this paper. 

Given this notation, we can now formulate the CART splitting rule as: 
$$ (j^*, s^*) = \argmin_{\substack{j \in \{1, \dots, p \}\\ s\in S_j}} {\cal L}(s),$$
and the chosen pair $ (j^*, s^*)$ is the split that will be carried out in practice. 

A naive implementation of this approach would require considering $\sum_j s_j$ possible splits, and performing $O(n)$ work for each one. In practice, the best split for each numerical/ordinal variable can be found in $O(n \log(n))$ operations, and for each categorical variable in $O(\max(n,  K_j  \log(K_j)))$ operations, for both regression and classification problems \citep{breiman1984classification,hastie2009elements}. Given these $p$ best splits, all that is left is to find their optimum. 

\subsection{LOO splitting rules}

For LOO splitting, our goal is to find the best variable to split on based on LOO scores for each variable under the given impurity criterion. The actual split performed is then the best regular CART split on the  selected variable. The resulting LOO splitting algorithm is presented in Algorithm \ref{alg:high-level LOO approach}.

\begin{algorithm}[H] 
\caption{High level ALOOF approach}
\begin{algorithmic} [1]
\FORALL{$j=1$ to $p$}
\STATE  $L(j)=0$
\FORALL{$i=1$ to $n$}
\STATE $s_{ij} = \arg\min_{s \in S_j} {\cal L}^{(-i)}(s)$
\STATE  $L(j) = L(j) + R(s_{ij},i)$
\ENDFOR
\ENDFOR
\STATE  $j* = \arg\min_{j \in \{1...,p\}} L(j)$
\STATE  $s* = \arg\min_{s \in S_{j*}} {\cal L} (s)$

\end{algorithmic}
\label{alg:high-level LOO approach}
\end{algorithm}

The result of Algorithm \ref{alg:high-level LOO approach} is again a pair $(j^*,s^*)$ describing the chosen split. 
The additional notations used in this description are: 
\begin{enumerate}
\item ${\cal L}^{(-i)}(s)$: the impurity criterion for candidate split $s$, when excluding the observation $i$. For example, if we assume WLOG that $i\in L(s)$ and the impurity criterion is the suaqred loss, we get 
$${\cal L}^{(-i)}(s) = \sum_{l \in L(s), l\neq i} (y_l - \bar{y}^{(-i)}_L)^2 +  \sum_{l \in R(s)} (y_l - \bar{y}_R)^2,$$
where $\bar{y}^{(-i)}_L$ is the average calculated without the $i$th observation. If the criterion is Gini or the observation falls on the right side, the obvious modifications apply. Consequently, $s_{ij}$ is the best split for variable $j$ when excluding observation $i$.  
\item $R(s,i)$: the LOO loss of the $i$th observation for the split $s$. In the above case (squared loss, left side) we would have: 
$$ R(s,i) = (y_i - \bar{y}^{(-i)}_L)^2,$$
with obvious modifications for the other cases. $L(j)$ is simply the sum of $R(s_{ij},i)$ over $i$, i.e., the total LOO loss for the $j$th variable when the best LOO split $s_{ij}$ is chosen for each $i$. 
\end{enumerate}

A naive implementation of this algorithm as described here would require performing $n$ times the work of regular CART splitting, limiting the usefulness of the LOO approach for large data sets. As we show below, for two-class classification we can almost completely eliminate the additional computational burden compared to regular CART splitting, by appropriately enumerating the LOO splits. For regression, the worst case performance of our approach is unfortunately close to being $n$ times worse than regular CART splitting. In practice, the added burden is much smaller.

\subsection{What is ALOOF estimating?} \label{unbias}

The quantity $L(j)$ in our LOO approach directly estimates the generalization error of splitting on variable $j$ using a CART impurity criterion: for each observation $i$, the best split is chosen based on the other $n-1$ observations, and judged on the (left out) $i^{th}$ observation. Hence $L(j)$ is an unbiased estimate of generalization impurity error for a split on variable $j$ based on a random sample of $n-1$ observations. The selected splitting variable $j^*$ is the one that gives the lowest unbiased estimate. Hence our goal of judging all variables in a fair manner,  that identifies the variables that are truly useful for reducing impurity and not just over-fitting is attained. 

Since we eventually perform the best split on our complete data at the current node ($n$ observations), there is still a small gap remaining between the ``$n-1$ observations splits'' being judged and the ``$n$ observations split'' ultimately performed, but this difference is negligible in most practical situations (specifically, when $n$ is large). 

\subsection{ALOOF stopping rules} \label{stop}

Tree methods like CART usually adopt a grow-then-prune approach, where a large tree is first grown, then pruned (cut down) to ``optimal'' size based on cross-validation or other criteria \citep{breiman1984classification}. 
Because cross-validation is built-in to the ALOOF splitting variable selection approach, it obviates the need to take this approach. For every variable, ALOOF generates an unbiased estimate of the generalization error (in Gini/squared loss) of splitting on this variable. If an additional split on any variable would be detrimental to generalization error performance, this would manifest in the ALOOF estimates (in expectation, at least) and the splitting would stop. This approach is not perfect, because ALOOF generates an almost-unbiased estimate for every variable, then takes the best variable, meaning it may still be over-optimistic and continue splitting when it is in fact slightly overfitting. However in our experiments below it is demonstrated that the combination of ALOOF's variable selection with ALOOF's stopping rules is superior to the CART approach. 

It should be noted that in Boosting or Random Forest approaches utilizing trees as a sub-routine, it is usually customary to avoid pruning and determine in advance the desirable tree size (smaller in Boosting, larger in Random Forest). Similarly, when we implement ALOOF within these methods, we also build the tree to the same desirable size, rather than utilize ALOOF stopping rules. 

\section{Efficient implementation} \label{Sec3}
As described above, the ALOOF algorithm utilizes a seemingly computationally expensive LOO cross-validation approach. However, we show that with a careful design, and under reasonable assumptions, we  are able to maintain the same computational complexity of the CART algorithm for two-class classification modeling. In addition, we show that for regression tasks ALOOF carries some increase of the computational burden, replacing a $O(log(n))$ factor in CART splitting rules with an $O(n)$ term.
We present our suggested implementation for both categorical and numerical/ordinal variables, in both two-class classification and regression settings. 

\subsection{Two-class classification --- categorical variables}

Assume a categorical variable with $K$ different categories. As described above, the CART algorithm performs a split selection by sorting the $K$ categories by their
mean response value and consider only splits along this sequence. Therefore, the complexity of finding the best split for this variable is $O(\max(K \log(K), n))$, where $n$ is the number of observations being split.

We now present our suggested implementation and compare its computational complexity with CART. 
First, we sort the observations according to the mean response value of the $K$ categories, just like CART does. Then, for each pair $(x_i,y_i)$ we leave out, we recalculate its category's mean response (an $O(1)$ operation), and resort the categories by their new means. Notice that since only a single category changes its mean ($x_i$'s category), resorting the sequence takes only $O(\log(K))$. Once the sequence is sorted again we simply go over the $K$ splits along the sequence. Therefore, for each pair $(x_i,y_i)$ ALOOF necessitates $O(K)$ operations. 
It is important to notice that despite the fact we have $n$ observations, there are only two possible pairs for each category, as $y_i$ takes only two possible values in the classification setting. This means we practically need to repeat the LOO procedure only for $2K$ pairs. Hence, our total computational complexity is $O(\max(K^2, n))$

We further show that under different reasonable assumptions, our computational complexity may even be equal to the complexity of CART splitting. 
First, assuming the number of categories grows no faster than the square root of $n$, $K^2=O(n)$, it is immediate that both algorithms result with a complexity of $O(n)$. This setting is quite common for the majority of categorical variables in real-world datasets. 

Second, under the assumption that a single observation cannot dramatically change the order of the categories in the sorted categories sequence, then the worst case complexity is lower than $O(\max(K^2, n))$. More specifically, assuming a single observation can move the position of its category by at most $B$ positions in the sorted categories sequence, then the complexity we achieve is $O(\max(K\log(K), KB, n))$. This assumption is valid, for example, when all categories have about the same number of observations. To emphasize this point, consider the case where the number of observations in each category is exactly the same, so that each category consist of $\sfrac{n}{K}$ observations. In this case the effect of changing the value of a single observation (on the category's mean) is proportional to ($\sfrac{K}{n}$). In other words, $\sfrac{K}{n}$ corresponds to the smallest possible difference between categories' means. Moreover, when leaving out a single observation, the change in this category's mean is bounded by this value. Hence, resorting the categories (following the LOO operation) is not more than an exchange of positions of adjacent categories in the sorted categories sequence ($B=1$). This leads to an overall complexity equal to CART's.

\subsection{Two-class classification --- numerical variables}
For numerical variables, CART performs a split selection by first sorting the $n$ pairs of observations according to their $x_i$ values, and then choosing a cut (from at most $n-1$ possible cuts) that minimizes the Gini criterion along both sides of the cut. By scanning along the list with sufficient statistics this takes $O(n)$ operations, leading to overall complexity $O(n\log(n))$ due to the sorting.

Our suggested implementation again utilizes the fact that $y_i$ takes only two possible values to achieve a complexity identical to CART's. Here we denote the values as $y_i \in \{-1,1\}$, 
we initialize our algorithm by sorting the observations in the same manner CART does. We then take out the observation $y_i=1$ with the lowest value of $x_i$, and find its best cut (by going over all $n-2$ possible cuts). We then place this observation back and take out the observations for which $y_i=1$, with the second lowest value of $x_i$. 
Notice that the only cuts whose Gini index values are affected by these operations are the cuts between the two pairs we replaced. Denote this group of cuts by $N_2$. This means that when exchanging the two observations we need to update the Gini values of at most $|N_2|$ cuts to find the best LOO cut. Continuing in the same manner leads to a total complexity of $O(\sum{|N_j|})=O(n)$ for finding the best cut of all observations with $y_i=1$. This process is then repeated with observations for which $y_i=-1$. Therefore, the overall complexity of our suggested implementation is $O(2n+n\log(n))=O(n\log(n))$, just like CART.
We present pseudo-code of our suggested approach in Algorithm \ref{alg:algorithm}. To avoid an overload of notation we only present the case of observations for which $y_i=1$. The key idea is presented in steps 20-21 of the algorithm, where it is emphasized that only the splits in $N_j$ have to be considered in every step.  

\begin{algorithm}\label{alg1}
\caption{LOO best split selection for numerical variables in two-class classification modeling}
\begin{algorithmic} [1]
\REQUIRE Observations $(x_1,y_1),\ldots,(x_n,y_n)$.
\REQUIRE $j=1$, $n_1=$number of observations for which $y_i=1$,  $optimal${\_}$cut=$ array of size $n_1$ holding the best LOO cut for each of the $y_i=1$ observations.
\STATE Sort observations according to $x_i$ values.

\STATE Set $A=\{(x_1,y_1),\ldots,(x_n,y_n)$ such that $y_i=1\}$.
\STATE Set $A=$sort $A$ according to its $x_i$ values in ascending order.

\STATE $A=A\setminus (x_1,y_1)$.
\STATE $optimal${\_}$cut_j=\infty$.
\STATE $optimal${\_}$cut${\_}$Gini=\infty$.

\FORALL{$(x_i, y_i)\in A $}
\STATE Set $cut_i=\frac{x_i+x_{i+1}}{2}$.
\STATE Set $Gini_i=$ Gini index value with respect to the cut $cut_i$.
\IF {$Gini_i<optimal${\_}$cut${\_}$Gini$}
\STATE $optimal${\_}$cut_j=cut_i$.
\STATE $optimal${\_}$cut${\_}$Gini=Gini_i$.
\ENDIF
\ENDFOR
\STATE $j=j+1$

\FORALL{$j\in\{2, \ldots, n_1\}$}
\STATE Set $A=A\cup (x_{j-1},y_{j-1})$.
\STATE Set $A=A\setminus(x_j,y_j)$.
\STATE $optimal${\_}$cut_j=optimal${\_}$cut_{j-1}$.
\STATE $N_j=\{(x_i,y_i) \in A$ such that $x_{j-1} <x_i<x_{j} \}$ .
\FORALL{$(x_i, y_i)\in N_j $}
\STATE Set $cut_i=\frac{x_i+x_{i+1}}{2}$.
\STATE Set $Gini_i=$ Gini index value with respect to the cut $cut_i$.
\IF {$Gini_i<optimal${\_}$cut${\_}$Gini$}
\STATE $optimal${\_}$cut_j=cut_i$.
\STATE $optimal${\_}$cut${\_}$Gini=Gini_i$.
\ENDIF
\ENDFOR
\STATE $j=j+1$.
\ENDFOR

\RETURN $optimal${\_}$cut$
\end{algorithmic}
\label{alg:algorithm}
\end{algorithm}

\subsection{Regression --- categorical variables}
The regression modeling analysis case is similar to the two-class classification in the presence of categorical variable, with a simple modification of the loss impurity criterion. 
Both CART and our suggested ALOOF method replace the Gini index with the squared error loss impurity criterion and follow the exact same procedures mentioned above. This leads to a overall complexity of $O(\max(K\log(K),n)$ for the CART algorithm and $O(\max(n,K^2))$ for our ALOOF method. 
As before, the computational complexity gap between the two methods may reduce under some reasonable assumptions. One simple example is when $K^2=O(n)$. Another example is in cases where we can assume that the effect of a single observation on the mean response value of each category is bounded. This happens, for instance, when the the range of the response values is bounded and there is approximately the same number of observations in each category.    

\subsection{Regression --- numerical variables}
In this setting as well, we compare both algorithms' implementation to those of the classification problem. 
The CART algorithm first sorts the $n$ observation pairs according to their $x_i$ values, and then chooses a cut that minimizes the squared error loss along both sides of the cut. As in the classification case, scanning along the list with sufficient statistics this takes $O(n)$ operations, leading to overall complexity $O(n\log(n))$ due to the sorting.

As we examine the LOO method in this regression setting, we notice that unlike CART, it fails to simply generalize the two-class classification problem. Specifically, it is easy to show that for each observation pair $(x_i, y_i)$ that is left out, the best cut may move in a non-monotonic manner (along the values of the numerical variable $x$), depending on the specific values of the observations that were drawn. This unfortunate non-monotonic behavior leads to a straight-forward implementation, where we first sort the observations by their $x_i$ values and then find the best cut for each observation that is taken-out by exhaustive search. The overall complexity is therefore $O(n^2)$, which compared with the CART implementation, replaces the $O(\log(n))$ factor with $O(n)$.

\section{Illustrative simulations} \label{Sec4}

\begin{figure}[!ht]
\centering
\includegraphics[width = 0.65\textwidth,bb= 30 100 750 490,clip]{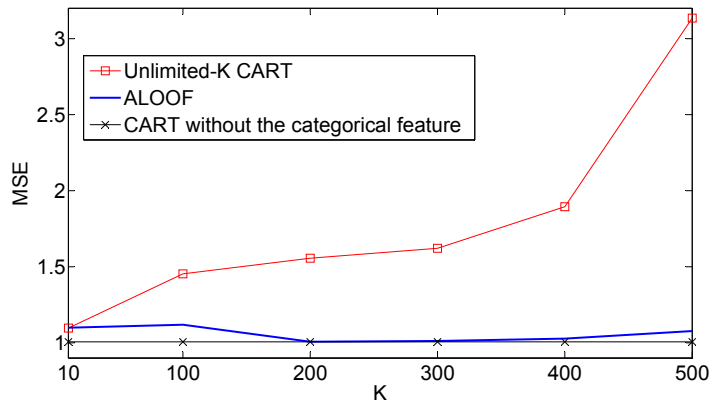}
\caption{Tree-based modeling in the presence of an uninformative categorical feature}
\label{fig:sim1_1}
\end{figure}

We start with two synthetic data experiments which demonstrate the behavior of ALOOF compared with CART in the presence of a categorical variable with varying number of categories. 
In the first experiment we examine our ability to detect a categorical variable which is not informative with respect to the response variable. Our goal is to model and predict the response variable from two different explanatory variables, where the first is numerical and informative while the other is categorical with $K$ different categories, and independent of the response variable.  We draw $1000$ observations which are split into $900$ observations for the train-set and $100$ for the test-set. We repeat this process multiple times and present averaged results. 
Figure \ref{fig:sim1_1} shows the test-set's average mean square error (MSE) for different number of categories $K$, using three different methods. The upper curve corresponds to the classic CART model, which is extremely vulnerable to categorical variables with large number of categories, as discussed above. The full line is our suggest ALOOF method, which manages to identify the uselessness of the categorical variable and discards it. The line with the {\sf x} symbols at the bottom is the CART model without the categorical variable at all, which constitutes a lower bound on the MSE we can achieve, using a tree-based model.   

\begin{figure}[!ht]
\centering
\includegraphics[width = 0.65\textwidth,bb= -10 100 710 490,clip]{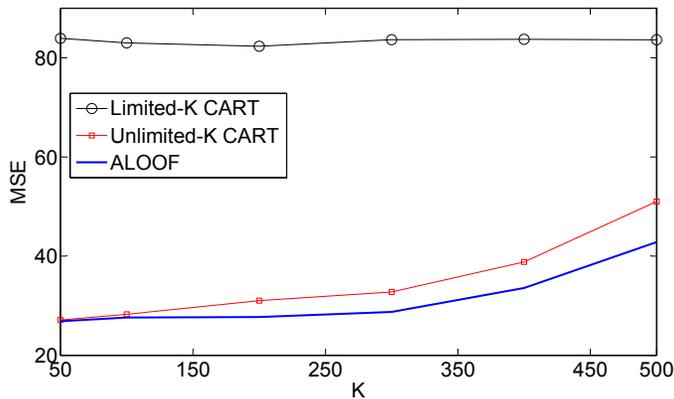}
\caption{Tree-based modeling in the presence of an informative categorical feature, (according to the model described in (\ref{intro_model}) with $\alpha=15$)}
\label{fig:sim1_2}
\end{figure}

In the second synthetic data experiment we revisit the model presented in the Introduction (\ref{intro_model}) as an example of a model with an informative categorical variable. We fix the strength of the interaction between the numerical and the categorical variables ($\alpha=15$) and repeat the experiment for different values of $K$. Figure \ref{fig:sim1_2} presents the results we achieve. As in the previous example ALOOF demonstrates superior performance to both its competitors as $K$ increases. This exemplifies the advantage of using ALOOF both when the categorical variable is informative or uninformative, for any number of categories $K$.  

We now turn to compare the modeling complexity of the ALOOF approach with that of CART. Since CART has a tree-depth parameter it may provide a variety of models which differ in their model complexity and the corresponding MSE they achieve. ALOOF, on the other hand, results in a single model, using the stopping policy described in Section \ref{stop}, with its corresponding MSE. Therefore, to have a fair comparison between the two methods we use degrees of freedom ($df$). We start with a brief review of the main ideas behind this concept. 

Following \cite{efron1986biased} and \cite{hastie2009elements}, assume 
the values of the feature vectors $X=(x_1,...,x_n)$  are fixed (the fixed-x assumption), and that the model gets one vector of response variable $Y=(y_1,...,y_n)$
for training, drawn according to the conditional probability model $p(Y|X)$ at the n data points. Denote by $Y^{new}$
another independent vector drawn according to the same distribution. $Y$ is
used for training a model $\hat{f}(x)$ and generating predictions $\hat{y_i}=\hat{f}(x_i)$ at the $n$ data points.
We define the training mean squared error as
\begin{equation}
MRSS=\frac{1}{n}\|Y-\hat{Y}\|^2_2
\end{equation}
and compare it to the expected error the same model incurs on the new, independent
copy, denoted in \cite{hastie2009elements} as $ERR_{in}$,
 \begin{equation}
ERR_{in}=\frac{1}{n}\mathbb{E}_{Y^{new}}\|Y^{new}-\hat{Y}\|^2_2
\end{equation}
The difference between the two is the \textit{optimism} of the prediction.
As Efron \citep{efron1986biased} and others have shown, the expected optimism in MRSS is
 \begin{equation}
\mathbb{E}_{Y,Y^{new}}(ERR_{in}-MRSS)=\frac{2}{n}\sum_{i}cov(y_i,\hat{y_i})\label{eq:ERR-MRSS}
\end{equation}
For linear regression with homoskedastic errors with variance $\sigma^2$, it is easy
to show that (\ref{eq:ERR-MRSS}) is equal to $\frac{2}{n}d\sigma^2$ where $d$ is the number of regressors,
hence the degrees of freedom. 
In nonparametric models (such as tree-based models), one usually cannot calculate the actual degrees of freedom of a modeling approach. However, in simulated examples it is possible to
generate good estimates $\hat{df}$ of $df$ through repeated generation of $Y, Y^{new}$ samples to empirically evaluate (\ref{eq:ERR-MRSS}). 

\begin{figure}[!ht]
\centering
\includegraphics[width = 0.65\textwidth,bb= 40 100 765 520,clip]{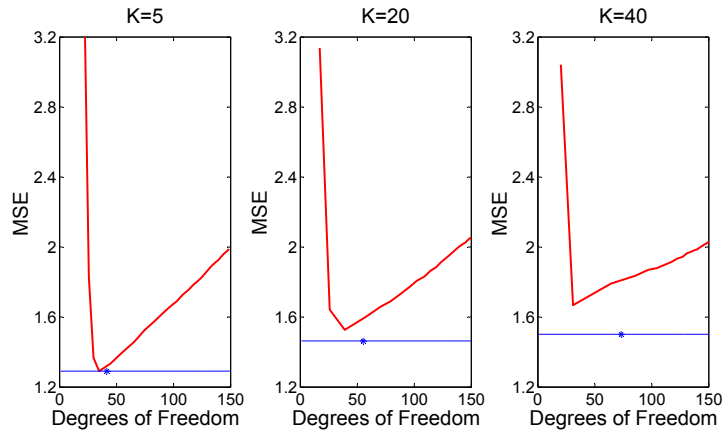}
\caption{Modeling complexity (represented in $df$) and its corresponding MSE in the presence of uninformative categorical feature with $K$ different categories. The curve corresponds to CART model, the asterisk is our ALOOF approach and the straight line is to emphasize that the MSE we achieve is uniformly lower than CART's curve}
\label{fig:sim2}
\end{figure}

In the following experiment we compare the models' degrees of freedom for CART and ALOOF. As explained above, CART generates a curve of values where the $df$ increases with the size of the tree (hence, the complexity of the model) and the MSE is the error it achieves on the test-set. Our ALOOF method provides a single value of MSE for the model we train. We draw $n=200$ observations from the same setup as in Figure \ref{fig:sim1_1} and achieve the results shown in Figure \ref{fig:sim2} for three different values of $K$. The straight line at the bottom of each figure is simply to emphasize that the MSE ALOOF achieves is uniformly lower than CART's curve, although it imposes more $df$ compared to the optimum of the CART curve. It is important to notice that this ``CART optimum" is based on Oracle knowledge; in practice, the pruned CART model is selected by cross-validation. Typically, it fails to find the model which minimizes the MSE on the test-set. 

As we look at the results we achieve for different values of $K$ we notice that ALOOF's superiority grows as the number of uninformative categories increases, as we expect.

\section{Real data study} \label{Sec5}

We now apply our ALOOF approach to real-world datasets and compare it with different commonly used tree-based methods. For obvious reasons, the data-sets we focus on are ones that include categorical features with a relatively large number of categories. All these dataset are collected from UCI repository \citep{Lichman:2013}, CMU Statlib \footnote{\url{http://lib.stat.cmu.edu/}} and Kaggle\footnote{\url{http://www.kaggle.com/competitions}} and are publicly available. 

Throughout this section we use ten-fold cross validation to achieve an averaged validation error on each dataset. In addition, we would like to statistically test the difference between ALOOF and its competitors and assert its statistical significance. However, the use of $K$-fold cross validation for model evaluation makes this task problematic, as the variance of cross validation scores cannot be estimated well \citep{bengio}. Some mitigations have been proposed to this problem in the literature, but we are not aware of a satisfactory solution.
Therefore we limit our significance testing to policies that circumvent this problem: 
\begin{itemize}
\item For larger datasets ($>1000$ observations) we use a single 90-10 training-test division to test for significance separately from the cross validation scheme. 
\item Combining the results of all datasets, we perform a sign test to verify the ``overall'' superiority of ALOOF.  
\end{itemize}

We start our demonstration with a comprehensive case study of the Melbourne grants dataset\footnote{\url{https://www.kaggle.com/c/unimelb}}

\subsection{Case study: Melbourne Grants} 
The Melbourne Grants dataset was featured in a competition on the Kaggle website. The goal was to predict the success of grant applications based on a large number of variables, characterizing the applicant and the application. The full training dataset, on which we concentrate, included 8708 observations and 252 variables, with a binary response (success/failure). We only kept 26  variables (the first ones, excluding some identifiers), and eliminated  observations that had missing values in these variables, leaving us with a dataset of $n=3650$ observations and $p=26$ variables. 
This dataset is interesting because some of the clearly relevant variables are categorical with many values, including: 
\begin{itemize}
\item The Field/Academic classification of the proposal (510 categories, encoded as codes from the Australian Standard Research Classification (ASRC)).
\item The Sponsor of the proposal (225 categories, an anonymized identifier).
\item The Department of the proposer (90 categories).
\item The Socio-economic code describing the potential impact of the proposed research (318 categories, encoded as codes from the Australian and New Zealand Standard Research Classification (ANZSRC)). 
\end{itemize}
Other clearly relevant variables include the history of success of failure by the current proposer, academic level, etc. 

\begin{figure}[htbp]
\centering
\includegraphics[width =\textwidth,bb= 80 70 715 550,clip]{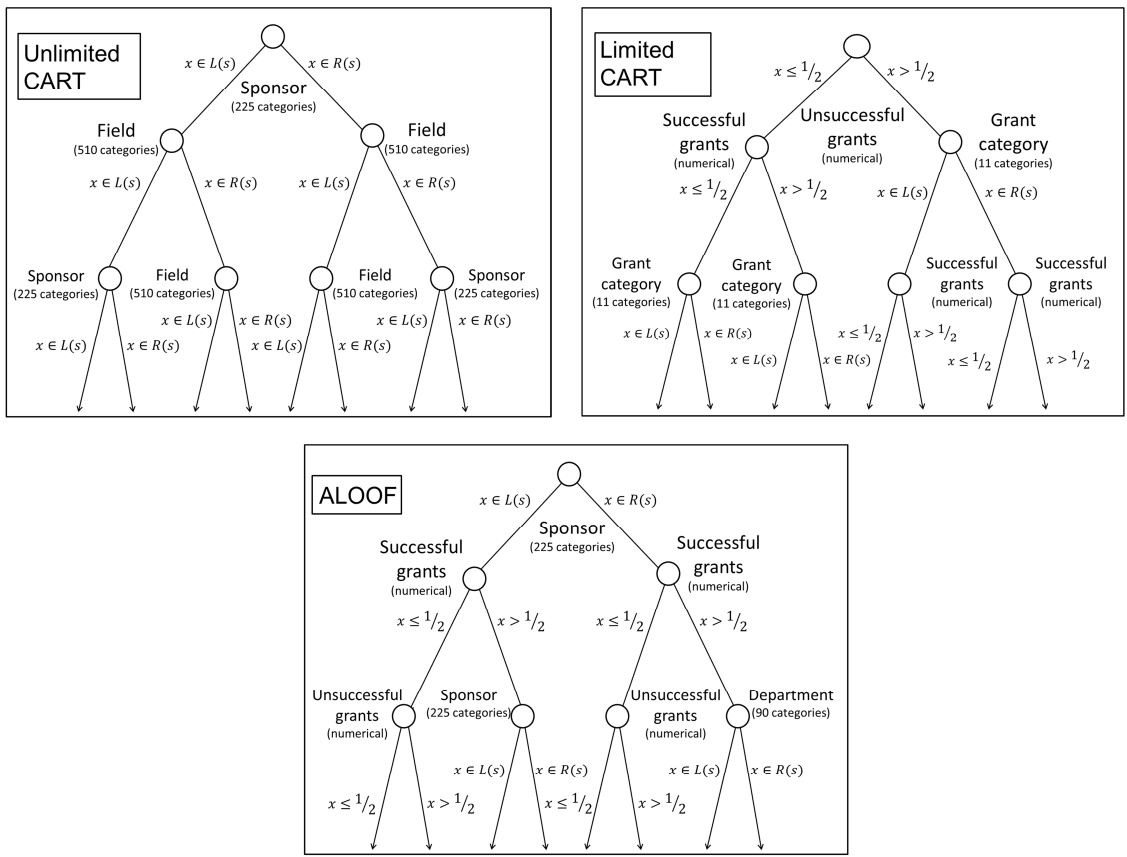}
\caption{Top three levels of trees built on the Melbourne Grants data using unlimited-K (standard) CART, limited-K CART (which eliminates categorical variables with $K>32$), and ALOOF.}
\label{fig:melb}
\end{figure}

In Figure \ref{fig:melb} we present the top three level of three trees built on the entire dataset, using unlimited-K CART, limited-K CART and ALOOF. As expected, the unlimited-K version uses exclusively the largest-K categorical variables, the limited-K version is not exposed to them, so identifies the history of success as the most relevant non-categorical information, while ALOOF combines the use of one categorical variable (Sponsor), with the history of success. The ten-fold cross-validation error of the three complete trees, as shown in Table \ref{table:melb}, is $0.245, 0.260, 0.218$, respectively. As discussed above, the ability to accurately infer significance from $K$-fold cross validation is limited. Therefore, in addition to the cross validation experiments, we arbitrarily partition the dataset into $90\%$  train-set and $10\%$ test-set, and apply a $t$-test on results we achieve on the test-set. 
This way we conclude that the advantage of ALOOF is statistically significant at level $0.05$ on a single random fold of the ten-fold CV.

The advantage of ALOOF over CART is preserved when they are used as sub-learners in Gradient Boosting (GB) and Random Forest (RF), also shown in Table \ref{table:melb}. For GB we use $50$ trees and limit their complexity by defining the minimum number of observations in the trees' terminal nodes to be $0.05\cdot n$. We train the model with a learning parameter of $\nu=0.1$ (see \citep{friedman2001} for further details). For the RF method we use {\sf Matlab}'s procedure {\sf treeBagger} with $500$ trees, where the rest of the parameters are maintained in their default values. Although the number of trees we utilize in both ensemble methods may seem relatively small, our experiments show that additional trees are not necessary as they do not significantly change the results we achieve.  

\begin{table}[htbp]
\caption{Misclassification results of two CART versions and ALOOF on the Melbourne Grant dataset. Results reported are from ten-fold CV. Results are shown for a single tree, Gradient Boosting trees and Random Forest (see text for details). The stars denote that the performance of ALOOF is statisticially significantly superior for all three versions.}
\centering
\renewcommand{\baselinestretch}{1}\footnotesize
\label{table:melb} 
\centering
\begin{tabular}{cccc}  
\hline
Approach & Unlimited-K & Limited-K & ALOOF\\
\hline
Single tree & $0.245$&$0.260$&$0.218^*$ \\
Gradient Boosting  &$0.186$ & $0.192$ &${0.165^*}$ \\
Random Forest &$0.172$ & $0.185$ &${0.153^*}$ \\
\hline
\end{tabular}
\end{table}

To gain further insight we can examine the variable importance measure commonly used in RF \citep{breiman2001random}, which judges the performance based on permutation tests on out-of-bag (OOB) samples. In Table \ref{tab:RFimp} we show the top five variables for each RF version. We can see that the lists of unlimited-K and limited-K CART have only one variable in common (Grant category code), while the ALOOF list combines the top three variables from the unlimited-K list and the top two from the limited-K list. This reiterates ALOOF's ability to take advantage of both categorical and numerical variables in a balanced manner.  

\begin{table}[htbp]
\caption{Top variables according to Random Forest's variable importance measure for each of the three tree-building approaches. As can be seen, ALOOF successfully identifies and combines the categorical and numeric variables that are most useful.}
\centering
\renewcommand{\baselinestretch}{1}\footnotesize
\label{tab:RFimp} 
\centering
\begin{tabular}{cccc}  
\hline
Rank & Unlimited-K & Limited-K & ALOOF\\
\hline
$1$ &  Sponsor&No. of successful grants & Sponsor \\
$2$  &Field & No. of unsuccessful grants &No. of successful grants \\
$3$  &Socio-economic code & Grant category code &No. of unsuccessful grants \\
$4$ &Department & Faculty & Field \\
$5$ & Grant category code & Grant's value & Socio-economic code \\ 
\hline
\end{tabular}
\end{table}

\subsection{Regression problems} 
We now turn to real-world regression problems. We first demonstrate our suggested approach with three classical small-size datasets.
In the Automobile dataset, we model the price of a car given a set of relevant features. This set of features include the car's brand which consists of $32$ categories. As previously discussed, the presence of such a categorical feature may cause overfitting as there are only $192$ observations. 
The Baseball experiment provides two designated datasets for hitters and catchers, where the goal is to predict the salary of each player in the preceding year. Here, the ``problematic" categorical feature is the player's team, consisting of $24$ different categories.

Table \ref{table:real-world-small} summarizes the results we achieve applying three different method. The CART column corresponds to a standard unlimited-K CART algorithm. CIT is the {\sf R} implementation of the Conditional Inference Tree ({\sf ctree}) using its default parameters of $10000$ Monte-Carlo replications to attain the distribution of the test statistic and a corresponding p-value threshold of $0.05$. ALOOF is our suggested algorithm.
As our datasets are relatively small, we use ten-fold cross validation to estimate an averaged $MSE$. Ours results demonstrate the advantage that ALOOF's LOO selection scheme has over both the overfitting approach of CART and significance-based approach of CIT in generating good prediction models. 

\begin{table}[ht]
\caption{Regression real-world data experiments on small datasets. For each dataset we introduce its categorical variables whose number of categories $K_j$ is relatively large. CART is a standard CART implementation, CIT corresponds to the Conditional Inference Trees \citep{hothorn2006unbiased} and ALOOF is our suggested method. The performance of each method is measured by averaged $MSE$, via ten-fold cross validation}
\centering
\renewcommand{\baselinestretch}{1}\footnotesize
\label{table:real-world-small} 
\centering
\begin{tabular}{ M{3.5cm} M{1.9cm} M{1.2cm} M{0.9cm} M{1.1cm} N}  

\hline

Dataset (size)
&\begin{tabular}{@{}c@{}} Categorical \\ variables ($K_j$) \end{tabular}
&CART
&CIT
&ALOOF&\\[15pt]

\hline

\begin{tabular}{@{}c@{}} Auto  $(192)$ \end{tabular}   
&\begin{tabular}{@{}c@{}} Car brand $(32)$ \end{tabular}  
&$9.48$
&$13.24$ 
&${8.22}$&\\[15pt]

\begin{tabular}{@{}c@{}} Baseball hitters $(264)$ \end{tabular}   
& Team $(24)$  
&$0.140$
&$0.136$ 
&$0.131$&\\[15pt]

\begin{tabular}{@{}c@{}} Baseball  catchers  $(176)$ \end{tabular}   
& Team $(24)$  
&$0.076$
&$0.075$ 
&$0.062$&\\[15pt]  

\hline

\end{tabular}
\end{table}
\vspace{1em}

In addition to these illustrative small-size datasets we apply our suggested method to more modern and large-scale regression problems. Table \ref{table:real-world-large-reg} presents the results we achieve in five different regression experiments. In Boston Housing we predict the value of each house, where the town's name is the categorical variable which may cause overfitting. In the Strikes experiment we model volume of large strikes (days lost due to industrial disputes) over a total of more than $30$ years. The name of the country in which the strike took place is the categorical variable with relatively many categories. The Donations dataset is from \textit{KDD} Cup of $1998$ (excluding missing values) where participants were asked to model the amount of donations raised, given a list of variables. These include several categorical variables with large number of variables, as described in Table \ref{table:real-world-large-reg}. The Internet usage dataset consists of two modeling problems. The first models the number of years a household is connected to an Internet provider while the second models the household's income. The variables used for modeling include multiple socio-economical variables where the problematic categorical ones are the occupation, language and the residence country of each household.

In all of these experiment we apply our ALOOF algorithm and compare it with a limited-$K$ and an unlimited-$K$ CART algorithms (as described above). Notice that in these large-scale experiments we cannot apply the CIT  method \citep{hothorn2006unbiased} as it is computationally infeasible.
In addition to these single tree models we also apply the ensemble methods Gradient Boosting (GB) and Random Forest (RF). We apply both of these methods using either unlimited-K CART trees or our suggested ALOOF method as sub-learners (limited-K is not competitive in relevant problems, where important categorical predictors are present, as demonstrated in the detailed Melbourne Grants case study). 
We apply GB and RF with the same set of parameters mentioned above. As in previous experiments we use ten-fold cross validation to achieve an averaged validation $MSE$. In addition, in order to attain statistically significant results we infer on a single random fold of the ten-fold CV for sufficiently large datasets (above $1000$ observations: the last three datasets in Table  \ref{table:real-world-large-reg}). A star indicates that the difference between ALOOF and its best competitor is statistically significant according to this $t$-test.

The results we achieve demonstrate the advantage of using ALOOF  both as a single tree and as a sub-learner in ensemble methods. Note that while the MSE ALOOF achieves in a single tree setting is consistently significantly lower than its competitors, it is not always the case with ensemble methods. This emphasizes the well-known advantage of using ensemble methods over a single tree, which can sometime mitigate the shortcomings of a single tree. However, it is still evident that ALOOF based ensemble methods are preferable (or at least equal) to the CART based methods. Moreover, it is notable that in some cases a single ALOOF tree achieves competitive results to ensemble based methods, such as in the Internet dataset.

\subsection{Classification problems}

In addition to the real-world regression experiments we also examine our suggested approach on a varity two-class classification problems. For each examined dataset we provide its number of observations and the portion of positive response values, as reference to the misclassification rate of the methods we apply. 
The Online Sales dataset describes the yearly online sales of different products. It is originally a regression problem which we converted to a two-class classification by comparing each response value to the mean of the response vector. Its variables' names are confidential but there exist several categorical variables with relatively large number of categories, as described in Table \ref{table:real-world-large-class}.  The Melbourne Grants dataset is described in detail above. The Price Up dataset provides a list of products and specifies whether the prices were raised at different stores in different time slots. It provides a set of variables which consist of two ``problematic" categorical variables: The name of the product and its brand. In the Salary Posting experiment our task is to determine the salary of a job posting given a set of relevant variables. As in the Online Sales, this dataset too is originally a regression problem which we converted to a two-class classification by comparing each response value with the mean of the response vector. Its ``large-$K$" categorical variables are the location of the job, the company and the title of the offered position. 
Lastly, the Cabs Cancellations dataset provides a list of cab calls and specifies which ones were eventually canceled.  The area code is the categorical variable which contains a large number of categories. 

As in the regression experiments, we apply different tree-based modeling methods to these datasets and compare the misclassification rate we achieve. In addition to the methods used in the regression problems we also apply the  method of \cite{sabato2008ranking}, which is designed for classification problems. We again perform a ten-fold cross validation to estimate the misclassification rate. As before, we also arbitrarily partition large datasets (the last four in Table \ref{table:real-world-large-reg}) into $90\%$  train-set and $10\%$ test-set, and apply a paired test on the difference between the results we achieve using ALOOF and each of its competitors. A star indicates that the difference between ALOOF and its best competitor is statistically significant.

As in the regression experiments, it is evident that our suggested ALOOF method is preferable both as a single tree and as sub-learner in ensemble methods. Note that in several experiments (such as Salary Postings) our advantage is even more remarkable compared with a limited-$K$ CART, which simply discards useful categorical variables only because they have too many categories. It is also notable that the single ALOOF tree is competitive with CART-based ensemble methods for several of the datasets. In particular, for the last three datasets the gain from replacing a single ALOOF tree with either GB or RF using CART is not statistically significant (whereas GB/RF + ALOOF does significantly improve on the CART ensemble counterparts).

In addition to the misclassification rate, we also evaluated the performance of our suggested algorithm under the Area Under the Curve (AUC) criterion. The results we achieved were very simialr with those in Table \ref{table:real-world-large-reg} and are omitted from this paper for brevity.

In order to obtain a valid overall statistical inference on the results we achieve, we examined the null hypothesis that ALOOF performs equally well to each of its competitors, based on all the datasets together, using a one-sided sign test as suggested by \citep{demvsar2006statistical}. 
For every competitor, we count the number of datasets (out of ten in Tables \ref{table:real-world-large-reg} and \ref{table:real-world-large-class} combined) in which ALOOF wins and apply a one-sided sign test. ALOOF outperforms both versions of CART on all datasets examined, performs better in GB for nine of ten, and for RF in eight of ten, with corresponding one-sided p values of 0.001, 0.01 and 0.05, respectively, after continuity correction.

\section{Discussion and conclusion}

In this paper we have demonstrated that the simple cross-validation scheme underlying ALOOF can alleviate a major problems of tree-based methods that are not able to properly utilize categorical features with large number of categories in predictive modeling. By adopting a LOO framework for selecting the splitting variable we allow ALOOF to eliminate categorical features that do not improve prediction, and select the useful ones even if they have many categories. 

As our simulations and real data examples demonstrate, the effect of using the ALOOF approach for splitting feature selection can be dramatic in improving predictive performance, specifically when categorical features with many values are present and carry real information (i.e., cannot simply be discarded). 

A key aspect of our approach is the design of efficient algorithms for implementing ALOOF that minimize the extra computational work compared to simple CART splitting rules. As our results in Section \ref{Sec3} indicate, ALOOF is more efficient in two-class classification than in regression. In practice, in our biggest regression examples in Section \ref{Sec5} (about $10,000$ observations), building an ALOOF tree takes up to $15$ times longer than building a similar size CART tree (the exact ratio obviously depends on other parameters as well). This means that at this data size, regression tree with ALOOF is still a practical approach even with GB or RF.  



For larger data sizes, an obvious mitigating approach to the inefficiency of ALOOF in regression is to avoid using LOO in this case, instead using L-fold cross validation with $L<<n$. This would guarantee the computational complexity is no larger than $L$ times that of CART splitting even with a naive implementation. The price would be a potential reduction in accuracy, as the {\em almost unbiasedness} discussed in Section \ref{unbias} relies on the LOO scheme. 

\section{Acknowledgments}
This research was partially supported by Israel Science Foundation grant 1487/12 and by a returning scientist fellowship from the Israeli Ministry of Immigration to Amichai Painsky. The authors thank Jerry Friedman and Liran Katzir for useful discussions. 

\begin{landscape}
\begin{table}
\caption{Regression real-world data experiments. For each dataset we introduce its categorical variables whose number of categories $K_j$ is relatively large. limited-$K$ CART is a CART implementation which discards categorical variables with $K_j>32$ while the unlimited-$K$ CART is an implementation which does not discard any variable. Gradient Boosting (GB) modeling with $50$ trees and Random Forest (RF) with $500$ are also provided, both with either CART or ALOOF as a sub-learner. The performance of each method is measured by averaged $MSE$, via ten-fold cross validation. In addition, a star indicates that the difference between ALOOF its best competitor in the same category (single tree, GB, RF) is statistically significant on a single arbitrary partitioning to train and test sets (this test was only applied to the last three datasets, with more than 1000 observations). }

\centering
\renewcommand{\baselinestretch}{1}\footnotesize
\label{table:real-world-large-reg} 
\centering
\begin{tabular}{M{1.6cm} M{2.6cm} M{1.5cm} M{1.75cm} M{1.1cm} M{1.2cm} M{1.2cm} M{1.2cm} M{1.2cm} N}  

\hline

Dataset (size)
&\begin{tabular}{@{}c@{}} Categorical \\ variables ($K_j$) \end{tabular}
&\begin{tabular}{@{}c@{}} Limited-$K$ \\  CART \end{tabular}
&\begin{tabular}{@{}c@{}} Unlimited-$K$ \\ CART \end{tabular}
&ALOOF
&\begin{tabular}{@{}c@{}} GB via \\ CART \end{tabular}
&\begin{tabular}{@{}c@{}} GB via \\  ALOOF \end{tabular}
&\begin{tabular}{@{}c@{}} RF via\\ CART \end{tabular}
&\begin{tabular}{@{}c@{}} RF via\\ ALOOF \end{tabular}&\\[20pt]

\hline

\begin{tabular}{@{}c@{}} Boston\\ Housing \\ $(506)$ \end{tabular}   
&Town name $(88)$ 
&$24.35$
&$24.07$ 
&${20.83}$
&$9.24$
&${7.88}$
&$9.08$
&${8.99}$& \\[30pt]

\begin{tabular}{@{}c@{}} Strikes \\ $(626)$ \end{tabular}   
&Country $(18)$ 
&$315.95$
&$315.95$ 
&${280.23}$
&$256.31$
&$257.12$
&$255.59$
&$256.06$& \\[25pt]

\begin{tabular}{@{}c@{}} Donations \\ $(6521)$ \end{tabular}   
&\begin{tabular}{@{}c@{}} Socio-cluster $(62)$, \\ ADI code $(180)$, \\ DMA code $(180)$, \\ MSA code $(245)$, \\ Mailing list $(348)$, \\  Zip code $(3824)$ 
 \end{tabular} 
&$30.99$
&$39.30$ 
&${22.92^*}$
&$23.34$
&${18.89^*}$
&$21.39$
&${16.28^*}$& \\[60pt]

\begin{tabular}{@{}c@{}} Internet -\\ years online \\ $(10108)$ \end{tabular}   
&\begin{tabular}{@{}c@{}} Occupation $(46)$, \\ Language $(99)$, \\  Country $(129)$  \end{tabular} 
&$0.727$
&$0.754$ 
&${0.703^*}$
&$0.699$
&$0.698$
&$0.698$
&$0.698$& \\[40pt]

\begin{tabular}{@{}c@{}} Internet -\\ household \\ income \\ $(10108)$ \end{tabular}   
&\begin{tabular}{@{}c@{}} Occupation $(46)$, \\ Language $(99)$, \\  Country $(129)$  \end{tabular} 
&$4.144$
&$4.308$ 
&${4.103^*}$
&$4.099$
&${4.005^*}$
&$3.991$
&$3.989$& \\[40pt]

\hline

\end{tabular}
\end{table}
\vspace{1em}

\end{landscape}

\begin{landscape}
\begin{table*}[htbp]
\caption{Classification real-world data experiments. The columns correspond to the same models as in the regression experiment, with the additional Sabato and Shalev-Shwartz (S\&S) algorithm \citep{sabato2008ranking} which is only applicable for classification problems. The performance of each method is measured by averaged misclassification rate, using ten-fold cross validation. In addition, a star indicates that the difference between ALOOF its best competitor in the same category (single tree, GB, RF) is statistically significant on a single arbitrary partitioning to train and test sets (this test was only applied to the last four datasets with more than 1000 observations).}
\centering
\renewcommand{\baselinestretch}{1}\footnotesize
\label{table:real-world-large-class} 
\centering
\begin{tabular}{ M{1.9cm} M{2.5cm} M{1.45cm} M{1.75cm} M{0.7cm} M{1.1cm} M{0.95cm} M{1.1cm} M{0.9cm} M{1.1cm} N}  

\hline

Dataset (size, portion of positives)
&\begin{tabular}{@{}c@{}} Categorical \\ variables ($K_j$) \end{tabular}
&\begin{tabular}{@{}c@{}} Limited-$K$ \\  CART \end{tabular}
&\begin{tabular}{@{}c@{}} Unlimited-$K$ \\ CART \end{tabular}
&S\&S
&ALOOF
&\begin{tabular}{@{}c@{}} GB via \\ CART \end{tabular}
&\begin{tabular}{@{}c@{}} GB via \\  ALOOF \end{tabular}
&\begin{tabular}{@{}c@{}} RF via\\ CART \end{tabular}
&\begin{tabular}{@{}c@{}} RF via\\ ALOOF \end{tabular}&\\[35pt]

\hline

\begin{tabular}{@{}c@{}} Online Sales \\ $(725, 0.29)$ \end{tabular}   
&\begin{tabular}{@{}c@{}} Confidential $(70)$ \\  Confidential $(273)$ \\Confidential $(406)$ \end{tabular}
&$0.198$
&$0.1903$ 
&$0.153$
&${0.151}$
&$0.145$
&${0.135}$
&$0.134$
&${0.121}$& \\[30pt]

\begin{tabular}{@{}c@{}} Melbourne \\ Grants \\ $(3650, 0.29)$ \end{tabular}   
&\begin{tabular}{@{}c@{}}Department $(99)$ \\ Sponsor $(234)$ \\  Field $(519)$ \end{tabular}
&$0.260$
&$0.245$ 
&$0.240$
&${0.218^*}$
&$0.186$
&${0.165^*}$
&$0.172$
&${0.153^*}$& \\[30pt]

\begin{tabular}{@{}c@{}} Price Up  \\ $(3980, 0.045)$ \end{tabular}   
&\begin{tabular}{@{}c@{}}Brand $(27)$ \\ Product $(216)$ \end{tabular}
&$0.035$
&$0.047$ 
&$0.036$
&${0.033^*}$
&$0.039$
&${0.034^*}$
&$0.034$
&$0.034$& \\[30pt]

\begin{tabular}{@{}c@{}} Salary\\ Posting  \\ $(10000, 0.31)$ \end{tabular}   
&\begin{tabular}{@{}c@{}}Location $(390)$ \\ Company $(390)$ \\ Title $(8324)$ \end{tabular}
&$0.330$
&$0.149$ 
&$0.148$
&${0.142^*}$
&$0.142$
&$0.142$
&$0.140$
&${0.138^*}$& \\[30pt]

\begin{tabular}{@{}c@{}} Cabs \\Cancellations  \\ $(34293, 0.096)$ \end{tabular}   
&Area code $(586)$
&$0.093$
&$0.095$ 
&$0.088$
&${0.083^*}$
&$0.088$
&${0.083^*}$
&$0.083$
&$0.083$& \\[30pt]

\hline

\end{tabular}
\end{table*}
\vspace{1em}

\end{landscape}

%
\bibliographystyle{abbrv}
\bibliography{sigproc}

\end{document}